\documentclass{article} 
\usepackage{iclr2016_workshop,times}
\usepackage{url}
\usepackage{amsmath}
\usepackage{amsfonts}
\usepackage{graphicx}

\title{Scale Normalization}

\author{Henry Z.~Lo, Kevin Amaral, \& Wei Ding \\
Department of Computer Science\\
University of Massachusetts Boston\\
Boston, MA 02155, USA \\
\texttt{\{henryzlo,ding\}@cs.umb.edu, kevin.m.amaral@gmail.com}
}

\begin{document}

\maketitle

\begin{abstract}
One of the difficulties of training deep neural networks is caused by improper scaling between layers.  Scaling issues introduce exploding / gradient problems, and have typically been addressed by careful scale-preserving initialization.  We investigate the value of preserving scale, or isometry,  beyond the initial weights.  We propose two methods of maintaing isometry, one exact and one stochastic.  Preliminary experiments show that for both determinant and scale-normalization effectively speeds up learning.  Results suggest that isometry is important in the beginning of learning, and maintaining it leads to faster learning. 
\end{abstract}

\section{Introduction}
The goal of many initialization methods is to preserve the gradient signal as it goes backwards through each layer of a neural network \cite{glorot,he,saxe}.  In RNNs, not preserving this signal may lead to the well-known vanishing and exploding gradient problems \cite{hochreiter}.  In general, learning in neural nets is much faster when the composite scales of all layers remains near a constant of the problem  \cite{saxe}.  

Results from this line of work suggest that initially preserving scale is conducive to learning.  However, any update rule which does not enforce scale-preservation will violate this condition (isometry) after the first iteration.  Does preserving scale continue to speed up learning after the first iteration?

There is evidence for and against.  On one hand, if scale were preserved throughout all epochs, the network would fail to learn non-isometric projections.  However, there is circumstantial evidence for the benefit of preserving scale during training:

\begin{itemize}
\item The objective most common in autoencoders produces an approximately isometric matrix $W$, and thus implicitly preserves scale (singular values) \cite{ae_pca}.
\item Co-training with both unsupervised and supervised objectives leads to faster-learning and more generalizable networks \cite{ladder}.  At least some of this result is due to the scale-preserving effect of the unsupervised objective, which effectively regularizes the singular values of each weight matrix.
\end{itemize}

The contribution of this work is to investigate the utility of scale-preserving constraints.  We separate this from the other effects of the unsupervised objective by normalizing scale without optimizing reconstruction.  Preliminary results with two different methods indicate that at least for the first few iterations of training, scale normalization leads to faster learning.

\section{Preserving Scale}

The forward scale of a layer with weight matrix $W$ is its effect on the length of its input $x$:  $\frac{\|W^Tx\|_2}{\|x\|_2}$.  This is a function of the singular values of $W$ (if $x$ were a singular vector, it is scaled by the corresponding singular value).

The backward scale of a layer is $\frac{\|Wd\|_2}{\|d\|_2}$, where $d$ is the incoming gradient.  Just as $W^T$ is used to calculate the forward pass, $W$ is used to calculate the outcoming gradient for the backward pass.  The forward and backward scales are both functions of the singular values $\sigma$, which are invariant under transposition.  

The magnitude of the gradient at a given layer $i$ is the product of the original gradient's magnitude and the scales of all layers $l>i$.  If all these scales (singular values) are greater than one, we have exploding gradients; if we have less than 1, we have vanishing gradients.

\cite{saxe} suggests using orthogonal matrices (or rectangular matrices with unit singular values) to avoid the complications of scale.  This effectively makes all matrices non-scaling (though length can still be reduced when components of $x$ lies in the nullspace of $W^T$).  This orthonormal initialization scheme contrasts with the popular initializations in \cite{glorot} and \cite{he}, which are based on preserving the variance of the forward and backward passes.  The benefit of the singular value interpretation of scale over their interpretation are thus:
\begin{itemize}
\item Unites the notions of forward and backwards scales (both functions of singular values).
\item Yields a constructive method of creating non-scaling weight matrices (orthonormal init).
\item Scaling as a function of singular values relies on much less assumptions than scaling as variance preservation.
\end{itemize}

\section{Normalizing Scale}

Here we propose multiple ways to normalize scale \textit{during} learning.  The challenge is to preserve scale, while retaining what is learned after each update.

It should be noted that simply making $W$ unitary after each update (setting all singular values to 1) would decorrelate $W'x$ and destroy information contained in $W$ 
.  It may not make sense to have every neuron's output to be in the same range.  So while orthonormalizing $W$ works for initialization, it does not make sense to do it in the course of training.


\subsection{Determinant Normalization}

One method of preserving the scale of each layer is to set its \textit{pseudo-determinant} to one.  Just as the determinant of a matrix is a product of its eigenvalues, the pseudo-determinant is a product of a (not necessarily square) matrix's singular values ($\prod_i \sigma_i$).  It is an aggregate measure of the scales of $W$.

If our update rule on $W$ gives us a new weight matrix $W^*$, we can determinant-normalize $W$:
\begin{equation}
W \leftarrow \frac{W^*}{det(W^*)}
\end{equation}
where \textit{det} is the product of the singular values of $W^*$.  This sets the determinant of $W$ to be one.  Alternatively, determinant-normalziation sets the geometric mean of the scales to one, in a sense "centering" the singular values.

\begin{figure}
\includegraphics[width=\linewidth]{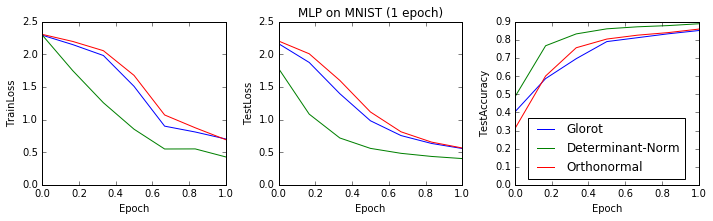}
\caption{Experiments over 1 epoch of training an MLP on MNIST.\label{mlp1}}
\end{figure}

\begin{figure}
\includegraphics[width=\linewidth]{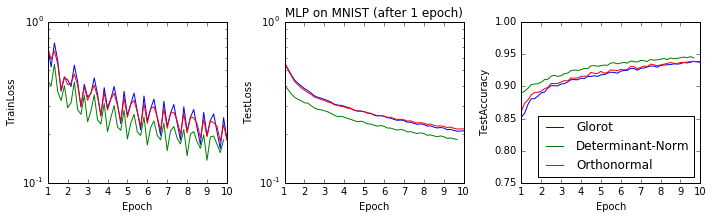}
\caption{Experiments over 10 epochs of training an MLP on MNIST. \label{mlp10}}
\end{figure}

We test determinant-normalization on a $100 \times 100 \times 100$ ReLu MLP on MNIST (see Figures \ref{mlp1} and \ref{mlp10}).  Models are trained using vanilla SGD, with batch size of 100.

Determinant normalization speeds up learning over Glorot and orthonormal initialization.  As expected, the benefits are most pronounced during the beginning of training (1 epoch is 500 batches).

\subsection{Scale Normalization}

Determinant normalization has the attractive property of being a function of $W$, not $x$.  However, calculating the pseudo-determinant runs into multiple problems:
\begin{itemize}
\item Calculating the pseudo-determinant requires an SVD to get the singular values, and is thus prohibitively expensive.
\item When singular values are less than 1 and matrices are large, it is very easy to get numerical underflow ($det(W^)=0$).
\end{itemize}

To address this, we propose an alternative notion of scale.  Empirically, the scaling of each vector $x$ by $W$ can be measured by the ratio:
\begin{eqnarray}
s = \frac{\|W^Tx\|_2}{\|x\|_2}
\end{eqnarray}

Thus, we can scale normalize by dividing $W$ by the average scale observed over a mini-batch:
\begin{equation}
W \leftarrow \frac{W^*}{E(s)}
\end{equation}

This sets the expected scaling to be one, such that on average, $W$ will preserve scale.  As there are many singular values, $s$ is different depending on $x$, and thus may be noisy depending on the batch, especially for small batches. 

\begin{figure}
\includegraphics[width=\linewidth]{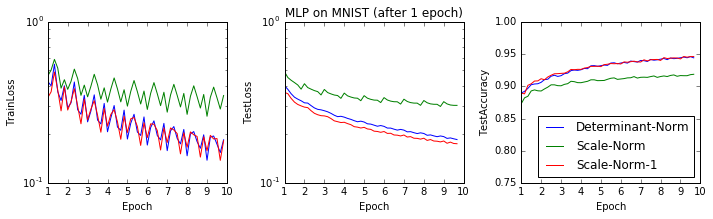}
\caption{Experiments over 10 epochs of training an MLP on MNIST.  Scale-Norm-1 indicates that scale normalization was only used for the first epoch.\label{mlpnporm10}}
\end{figure}

We noted that scale-normalization does not help after the first epoch.  When running scale-normalization for only the first epoch (100 iterations), it achieves similar results to determinant normalization.

\section{Future Work}

Preliminary results indicate that maintaining isometry is useful for learning, at least in the beginning.  Future work will relate scale-normalization to batch-normalization, and more advanced optimization algorithms.  Experiments on larger datasets such as CIFAR10 and convolutional architectures are in progress.  We believe that investigating how isometry interplays with learning speed will bring insight into how to speed up learning in the future.

\bibliography{iclr2016_workshop}
\bibliographystyle{iclr2016_workshop}

\end{document}